\documentclass[sigplan]{acmart}
\usepackage{microtype}
\AtBeginDocument{%
  }

\acmConference[Agent Skills '26]{The First Workshop on Agent Skills,
  co-located with ACM CAIS 2026}{May 26, 2026}{San Jose, CA, USA}
\setcopyright{none}
\renewcommand\footnotetextcopyrightpermission[1]{}
\begin{document}

\title[When Skills Don't Help]{When Skills Don't Help: A Negative Result on Procedural Knowledge for Tool-Grounded Agents in Offensive Cybersecurity}
\subtitle{Position Short Paper}

\author{Samuel Jacob Chacko}
\authornote{Both authors contributed equally to this research.} 
\affiliation{%
  \institution{Florida State University}
  \city{Tallahassee}
  \state{Florida}
  \country{USA}
}
\email{sj21j@fsu.edu}

\author{James Hugglestone}
\authornotemark[1]
\affiliation{%
  \institution{Florida State University}
  \city{Tallahassee}
  \state{Florida}
  \country{USA}
}
\email{jah21e@fsu.edu}

\author{Chashi Mahiul Islam}
\affiliation{%
  \institution{Florida State University}
  \city{Tallahassee}
  \state{Florida}
  \country{USA}
}
\email{cislam@fsu.edu}

\author{Xiuwen Liu}
\affiliation{%
  \institution{Florida State University}
  \city{Tallahassee}
  \state{Florida}
  \country{USA}
}
\email{xliu@fsu.edu}

\renewcommand{\shortauthors}{S. J. Chacko et al.}

\begin{abstract}
  Agent Skills, structured packages of procedural knowledge loaded into an LLM agent at inference time, are widely reported to improve task pass rates by an average of 16.2~percentage points across diverse domains. Yet the same benchmarks show wide variance, with 16 of 84 tasks suffering negative deltas when Skills are introduced. The community has not yet articulated a clean mechanism for \emph{when} Skills help and when they are merely redundant overhead. We re-analyze a recently published 180-run controlled study of an MCP-grounded autonomous Capture-the-Flag (CTF) agent under four documentation conditions of increasing richness (591, 12865, 17253, and 36001 tokens) and show that these conditions correspond almost exactly to a No-Skills, Experiential-Skills, Curated-Skills, and Comprehensive-Skills ablation. In offensive cybersecurity, a domain not deeply covered by existing Skills benchmarks, the marginal benefit of Skills collapses. The spread between the no-Skills and full-Skills conditions is only 8.9~pp ($p = 0.71$, $\chi^2$; $p = 0.25$, Cochran--Armitage trend test; five of six pairwise Cohen's $h$ values fall below the $0.2$ small-effect threshold). We argue that the missing variable is \emph{environment-feedback bandwidth}. When an agent's tool layer returns strict, schema-validated, low-latency observations, the environment itself supplies the procedural correction signal that Skills are normally needed to provide. As a result, the marginal benefit of curated Skills diminishes substantially, and, in some cases (e.g., our timing side-channel setting), actively degrades performance. We articulate a falsifiable hypothesis, sketch its design implications for compound AI systems, and will release the reanalysis pipeline to support replication.
\end{abstract}




\maketitle

\section{Introduction}

\label{sec:intro}

Agent Skills~\cite{anthropic-skills,agentskills-spec} have emerged as a unifying abstraction for the procedural knowledge that LLM agents consult at inference time. A Skill is a folder containing a \texttt{SKILL.md} file with metadata and instructions, optionally bundled with scripts and references, that an agent loads on demand. The format was released as an open standard in late 2025 and has since been adopted across more than thirty agent platforms~\cite{jiang2026skills-sok}, with community marketplaces hosting tens of thousands of community-authored Skills. The premise is intuitive: foundation models supply broad capability, and Skills supply the domain-specific recipes the model otherwise lacks.

The first systematic empirical evidence on whether Skills work at scale arrived with SkillsBench~\cite{li2026skillsbench}, a benchmark of 84 tasks across 11 domains, evaluated across 7 agent--model configurations and 7,308 trajectories. Two findings define the current state of the field. First, curated Skills add a substantial average of $+16.2$ percentage points (pp) to pass rates. Second, that average hides extreme variance: domain effects range from $+51.9$~pp (healthcare) down to $+4.5$~pp (software engineering), and 16 of 84 individual tasks show \emph{negative} deltas when Skills are provided. The authors of SkillsBench are explicit that the field lacks a mechanistic account of when Skills help, when they are inert, and when they degrade performance.

This paper offers one such account, grounded in a re-analysis of a recently published 180-run controlled study of an autonomous CTF agent~\cite{hugglestone2026striatum} that was designed and run before the \texttt{SKILL.md} specification was published, but whose four documentation conditions correspond, structurally and operationally, to a clean Skills ablation. The agent is built on the Model Context Protocol (MCP)~\cite{anthropic-mcp}, a schema-validated tool interface that returns strictly typed observations from a containerized arsenal of reverse-engineering, web exploitation, and binary-analysis tools. This schema-enforcement principle is analogous to grammar-constrained decoding~\cite{geng2023grammar}, which has been shown to substantially reduce hallucinated outputs in structured prediction. We argue that this MCP-grounded execution stack is the moderating variable that has been missing from Skills evaluation: rich, deterministic, low-latency tool feedback substitutes for a substantial fraction of the procedural guidance that curated Skills are normally needed to supply.

We make three contributions:

\begin{itemize}
\item \textbf{A new domain entry.} We extend the empirical landscape of   Skills evaluation into offensive cybersecurity (binary exploitation,  web exploitation, reverse engineering, cryptography), a domain underrepresented in current Skills benchmarks. Our re-analysis covers 180 trajectories across 15 challenges and a four-level documentation ablation.
\item \textbf{A null result at the sample size studied.} On this domain, structured procedural knowledge, spanning attack templates, distilled lessons, and a comprehensive Skills bundle, adds at most $+8.9$~pp over an MCP-only baseline (95\% CI for the difference: $[-6.8, +24.6]$~pp). None of the differences reach statistical significance ($p = 0.71$, $\chi^2$; $p = 0.25$, Cochran--Armitage trend test; five of six pairwise Cohen's $h$ values fall below the $0.2$ small-effect threshold, with No-Skills vs.\ Comprehensive at $h = 0.23$). We do not claim a causal demonstration of zero effect; we claim only that any effect in this domain is below the resolution of a 180-run controlled study. The Comprehensive-Skills condition is also outperformed by a Curated-Skills condition on at least one challenge class.
\item \textbf{The feedback-bandwidth hypothesis.} We articulate a falsifiable mechanistic hypothesis: \emph{the marginal benefit of Skills is inversely related to the bandwidth of deterministic environment feedback available to the agent.} We sketch the implications for compound AI system design and propose a concrete experimental program to test the hypothesis at scale.
\end{itemize}

The paper proceeds as follows. Section~\ref{sec:framing} positions the four documentation conditions of the source study as a Skills ablation. Section~\ref{sec:results} presents the re-analyzed numbers, including a case-study where the Comprehensive-Skills condition underperforms a narrower Curated-Skills condition. Section~\ref{sec:hypothesis} articulates the feedback-bandwidth hypothesis and its testable predictions. Section~\ref{sec:discussion} situates the result in the Skills literature and draws implications for practitioners building MCP-grounded agents. Section~\ref{sec:limits} addresses limitations.

\section{Reframing the Study as a Skills Ablation}
\label{sec:framing}

The source study~\cite{hugglestone2026striatum} evaluated an MCP-grounded LLM agent on 15 CTF challenges (memory corruption, reverse engineering, web exploitation, cryptography) under four conditions defined by the \emph{type} and \emph{volume} of context-attached procedural documentation. Each condition was run on every challenge with three independent trials, yielding $15 \times 4 \times 3 = 180$ trajectories. The model (Claude Sonnet 4.5 with extended thinking) and the tool layer (Nmap, Ghidra, Angr, GDB, exposed via MCP servers with strict JSON-schema validation) were held constant; only the procedural documentation varied. Importantly, each challenge is a long-horizon task~\cite{liu2024lost} that requires sequential tool invocation, hypothesis revision, and multi-phase exploitation. Trajectories average 17--20 minutes per run, making 15 challenges a substantive evaluation despite the modest count.

We map these conditions onto the standard Skills taxonomy:

\begin{itemize}
\item \textbf{No-Skills (Minimal, 55 lines).} Only MCP server definitions and tool schemas. No \texttt{SKILL.md}-like procedural guidance. This is the \emph{No-Skills} condition in the SkillsBench taxonomy~\cite{li2026skillsbench}.
\item \textbf{Experiential-Skills (Lessons, 1{,}478 lines).} Lessons distilled from prior debugging sessions (heap exploitation pitfalls, web-framework patterns, etc.) plus a triage classifier. Structurally, this is a self-distilled, trajectory-mined Skills bundle, procedural knowledge \emph{after the fact}.
\item \textbf{Curated-Skills (Templates, 1{,}976 lines).} A structured multi-phase solving protocol with attack templates indexed by vulnerability type. This is the \emph{Curated-Skills} condition: focused, human-authored procedural knowledge with explicit applicability conditions and parameterized action sequences, exactly what the skill file (\texttt{SKILL.md}) specification anticipates.
\item \textbf{Comprehensive-Skills (Baseline, 4{,}147 lines).} The union of Experiential and Curated, plus a set of orchestration and meta-documentation files (\texttt{START\_HERE.md}, \texttt{INDEX.md}, \texttt{HOW\_TO\_USE.md}, \texttt{SYSTEM\_SUMMARY.md}, and related navigation scaffolding) totaling approximately 693 additional lines. The triage tool itself is present in all augmented conditions (Experiential, Curated, and Comprehensive); what is unique to Comprehensive is the orchestration layer. This is the current best-practice ``everything we have'' Skills configuration.
\end{itemize}

The framing is structurally analogous rather than literal. The source study's conditions were not authored as \texttt{SKILL.md} files and do not implement the on-demand loading semantics of the formal specification, in which an agent first reads a Skill's name and description and then selectively loads full bodies as needed. Our conditions front-load all procedural content at task initialization, which inflates context cost relative to a spec-compliant Skills deployment. What the conditions do share with Skills is their functional role: each supplies a different volume and type of procedural knowledge, layered over an identical model and tool stack, with the Minimal condition acting as the No-Skills control. With that caveat in mind, the study is, to our knowledge, the largest controlled procedural-knowledge ablation in offensive cybersecurity to date, and the only one whose underlying environment is fully MCP-grounded.

\section{Results}
\label{sec:results}

Table~\ref{tab:results} reports task pass rates and mean solve times under each Skills condition, alongside SkillsBench's reported domain-wise gains for context.

\begin{table}[ht]
\caption{Pass rates and mean solve times across Skills conditions on the 15-challenge offensive-security benchmark (3 trials per challenge per condition; 180 runs total). $\Delta$ vs. No-Skills shown in pp. For reference, SkillsBench~\cite{li2026skillsbench} reports per-domain
  $\Delta$ ranging from $+4.5$~pp (software engineering) to $+51.9$~pp (healthcare), with an aggregate of $+16.2$~pp.}
\label{tab:results}
\resizebox{\linewidth}{!}{%
\begin{tabular}{@{}lrrrrr@{}}
\toprule
Condition & Lines & Tokens & Pass Rate & $\Delta$ (pp) & Mean Time \\
\midrule
No-Skills (Minimal)        & 55      & 591      & 77.8\% (35/45) & --    & 20.1 min \\
Experiential (Lessons)     & 1{,}478 & 12{,}865 & 82.2\% (37/45) & +4.4  & 19.1 min \\
Curated (Templates)        & 1{,}976 & 17{,}253 & 84.4\% (38/45) & +6.6  & 18.5 min \\
Comprehensive (Baseline)   & 4{,}147 & 36{,}001 & 86.7\% (39/45) & +8.9  & 17.1 min \\
\bottomrule
\end{tabular}%
}
\end{table}

\begin{figure}[t]
\centering
\includegraphics[width=\columnwidth]{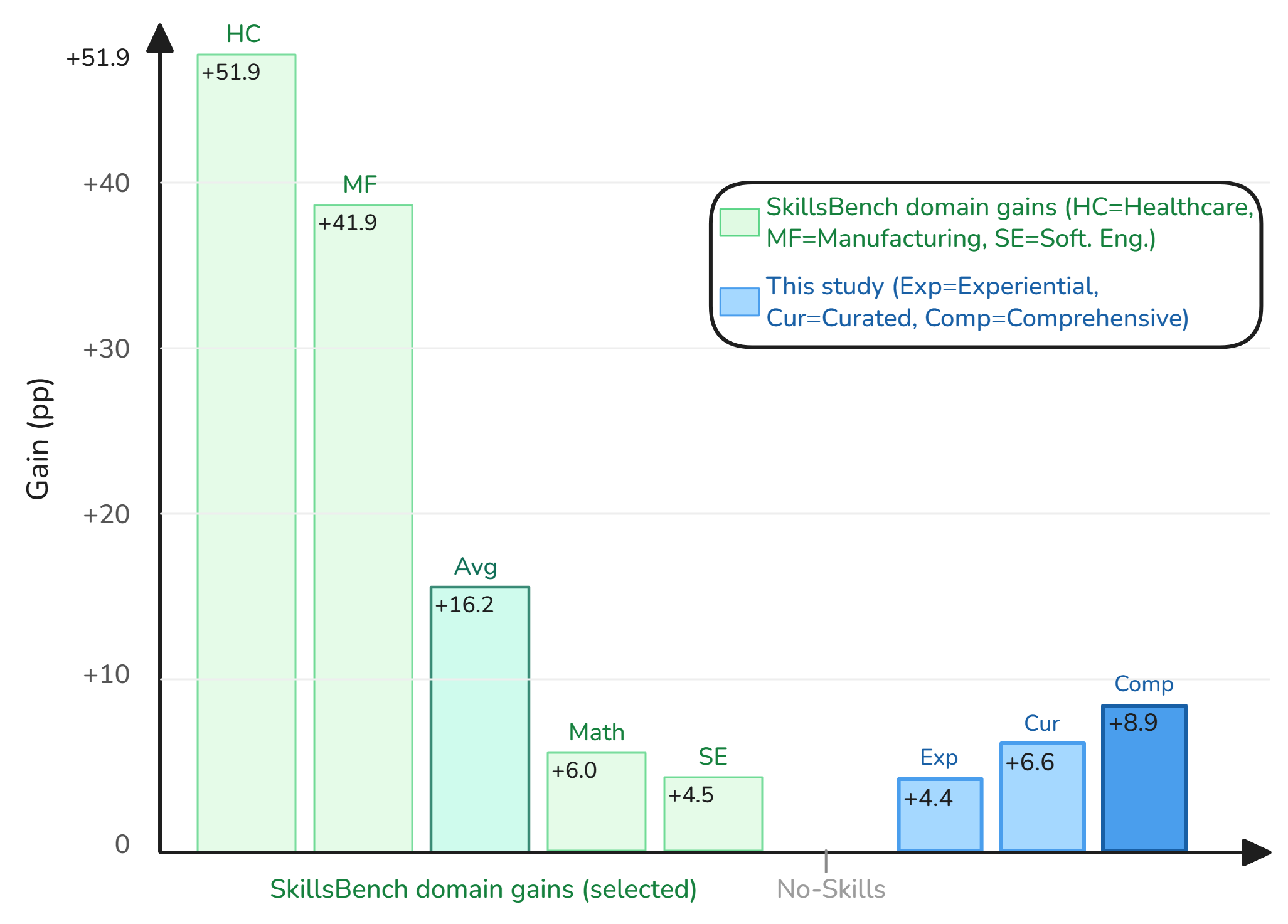}
\caption{Skills gain (pp above no-Skills baseline) by condition. Blue bars: this study's four documentation conditions on the 15-challenge offensive-security benchmark ($p = 0.71$, $\chi^2$; $p = 0.25$, Cochran--Armitage trend test; five of six pairwise Cohen's $h$ values below $0.2$). Green bars: selected SkillsBench domain gains for reference~\cite{li2026skillsbench} (HC\,=\,Healthcare $+51.9$\,pp; MF\,=\,Manufacturing $+41.9$\,pp; Avg\,=\,cross-domain mean $+16.2$\,pp; SE\,=\,Software Engineering $+4.5$\,pp). Our Comprehensive condition ($+8.9$\,pp) sits within the range of SkillsBench's lowest-gain domain and well below the cross-domain average, consistent with the feedback-bandwidth hypothesis (Section~\ref{sec:hypothesis}).}
\Description{A bar chart comparing Skills gain in percentage points above a no-Skills baseline. Two groups of bars are shown: green bars representing selected SkillsBench domain gains (Healthcare +51.9 pp, Manufacturing +41.9 pp, cross-domain average +16.2 pp, Software Engineering +4.5 pp), and blue bars representing this study's four conditions on a 15-challenge offensive cybersecurity benchmark (Experiential +4.4 pp, Curated +6.6 pp, Comprehensive +8.9 pp). The blue bars are clustered at the low end of the green reference bars, with all differences between the blue conditions being statistically non-significant.}
\label{fig:gain}
\end{figure}

\paragraph{Headline number.} The Comprehensive-Skills condition adds $+8.9$~pp over the No-Skills baseline (95\% CI for the difference: $[-6.8, +24.6]$~pp, normal approximation for the difference of two proportions). This interval excludes the SkillsBench healthcare ($+51.9$~pp) and manufacturing ($+41.9$~pp) high-gain domains but contains both the cross-domain mean ($+16.2$~pp) and the low-gain software-engineering domain ($+4.5$~pp). We can therefore distinguish our point estimate from the highest-gain SkillsBench domains but not from its overall mean; the comparison in Figure~\ref{fig:gain} should be read as a comparison of point estimates rather than a formal between-study test.

\paragraph{Statistical significance.} A $\chi^2$ test of independence on the $4 \times 2$ outcome table (condition $\times$ success/failure) returns $p = 0.71$, far above any conventional threshold. Recognizing the natural ordering of conditions by procedural-content volume, a Cochran--Armitage trend test yields $Z = 1.15$, $p = 0.25$---directionally consistent with a monotonic effect but well short of significance at this sample size. A Kruskal--Wallis test on the solve-duration distributions returns $p = 0.77$. Pairwise Cohen's $h$ values (the appropriate effect size for proportions) range from $0.06$ to $0.23$; five of six pairwise comparisons fall below the conventional $0.2$ small-effect threshold, with only No-Skills vs.\ Comprehensive ($h = 0.23$) at the boundary of a small effect. Under any reasonable inferential standard, \emph{Skills did not significantly improve performance in this study}.

\paragraph{The non-monotonicity case.} On the timing-side-channel sub-task, the rich-Skills conditions did \emph{not} dominate the focused ones. The Comprehensive-Skills condition succeeded on 1 of 3 trials ($33\%$); the Curated-Skills condition succeeded on 2 of 3 ($67\%$); the No-Skills condition failed all 3 ($0\%$). Trajectory inspection in the source study indicates that the \emph{additional} procedural content in Comprehensive (the experiential lessons) biased the agent toward symbolic execution. The source study~\cite{hugglestone2026striatum} characterizes this as \emph{false lesson propagation}, where the agent encodes a spurious causal lesson from a small number of prior trajectories and applies it in an inappropriate context. In this case, the transferred technique is not suited to side-channel attacks and is notably absent from the narrower Templates condition. We treat this as an illustrative case rather than a robust effect, given the small per-condition sample ($n=3$) at the sub-task level. With that caveat, the pattern aligns with the ``negative-delta'' phenomenon SkillsBench reports for 16 of 84 of its tasks~\cite{li2026skillsbench}: more procedural knowledge can reduce performance when it crowds out the right move.

\paragraph{Token economy.} The Comprehensive-Skills condition consumed roughly $61\times$ more procedural-context tokens than the No-Skills condition (36{,}001 vs.\ 591 tokens; 4{,}147 vs.\ 55 lines) for a non-significant $+8.9$~pp gain. We report both line and token counts because the two ratios differ: lines reflect authoring effort, but tokens are what the model pays for at inference. Crucially, the practical argument does not depend on statistical significance: even accepting the observed $+8.9$~pp at face value, the $\sim$60$\times$ token overhead makes the No-Skills condition the rational engineering choice in this domain. Under any cost-aware deployment metric, the No-Skills condition is on the Pareto frontier.

\section{The Feedback-Bandwidth Hypothesis}
\label{sec:hypothesis}

We propose the following hypothesis as an explanation:

\begin{quote}
\textbf{H1 (Feedback-Bandwidth).} The marginal benefit of curated Agent Skills is inversely related to the bandwidth of deterministic environment feedback available to the agent during task execution. In environments with high feedback bandwidth, the environment itself supplies the procedural correction signal that Skills normally provide, and the marginal benefit of curated Skills diminishes substantially, potentially to within statistical noise.
\end{quote}

\paragraph{Operationalization.} We define an environment's \emph{feedback bandwidth} along three axes: (i) \emph{determinism}, whether identical actions yield identical observations; (ii) \emph{schema fidelity}, whether observations are typed and structured rather than unstructured natural language; and (iii) \emph{latency}, whether feedback arrives in a timescale that permits in-task corrective replanning. CTF challenges mediated through MCP score high on all three~\cite{geng2023grammar}: \texttt{nmap} returns typed port lists, \texttt{gdb} \texttt{inspect\_heap} returns structured JSON, and exploit attempts succeed or fail against a binary verifier (the flag) within seconds. By contrast, the SkillsBench domains where Skills are most beneficial, healthcare ($+51.9$~pp), manufacturing ($+41.9$~pp)~\cite{li2026skillsbench}, are precisely those where the ``environment'' is brittle, partially observable formats with weak, delayed, or absent corrective feedback.

\paragraph{Predictions.} H1 makes three testable predictions. 
\begin{itemize}
    \item \textbf{P1:} Within a fixed domain, lowering feedback bandwidth (e.g., replacing schema-validated tool returns with raw shell output) should increase the marginal benefit of Skills.
    \item \textbf{P2:} Within a fixed environment, tasks whose verifiers are dense and immediate should show smaller Skills deltas than tasks whose verifiers are sparse and delayed.
    \item \textbf{P3:} Adding procedural knowledge that contradicts environment feedback should produce negative deltas, while adding procedural knowledge that the environment cannot supply (e.g., domain conventions) should produce positive deltas, even within the same task suite.
\end{itemize}  
None of these predictions are tested by existing Skills benchmarks; all are tractable in 2026 with current agent harnesses~\cite{li2026skillsbench}.

\paragraph{Design Implications.} If H1 holds, the implicit practitioner advice in the Skills literature should be reframed as domain-conditional rather than universal, though we note this prescription rests on a single-domain, single-model study and is intended as a hypothesis to test rather than a deployment recommendation. The current implicit recommendation is ``invest in curated Skills first, since they yield $\sim 16$~pp.'' Under H1, the recommendation becomes domain-conditional: if your environment can support rich, deterministic, low-latency tool feedback (a strong MCP implementation, structured tool outputs, fast verifiers), invest there first; Skills are a smaller follow-on lever. If the environment cannot support this, then in domains like healthcare, document workflows, and multi-system enterprise tasks, curated Skills become the dominant lever. This reframes Skills from a universally beneficial layer into a \emph{compensatory} layer whose value is determined by the environment it sits over.

\section{Discussion}
\label{sec:discussion}

\paragraph{Relation to SkillsBench.} Our result complements rather than contradicts SkillsBench~\cite{li2026skillsbench}. SkillsBench reports a positive aggregate effect across 11 domains and acknowledges a long tail of domains and tasks where Skills are inert or harmful. We supply the missing end of the distribution: a high-feedback-bandwidth domain where the effect collapses to non-significance, and a candidate mechanism for the long-tail heterogeneity. The two findings together suggest a research program in which the Skills benefit is decomposed by environment-feedback profile, rather than reported as a single domain-level number.

\paragraph{Relation to self-generated Skills.} SkillsBench also reports that self-generated Skills produce no net benefit ($-1.3$~pp on average). Our Experiential condition is structurally analogous: procedural knowledge is distilled \emph{post hoc} from prior trajectories rather than authored upfront, and it shows a similarly small absolute effect ($+4.4$~pp, not significant). This is consistent with the broader observation that LLM-authored procedural knowledge provides weaker grounding than direct environmental feedback. Under H1, this convergence is not incidental: self-distilled procedural knowledge is, in effect, a compressed and lossy summary of the very feedback signal it was generated from. When the agent has access to that feedback at task time, the distilled Skills are largely redundant; the $-1.3$~pp aggregate effect in SkillsBench and the $+4.4$~pp non-significant effect here both fit this reading. Self-generated Skills are therefore the cleanest test bed for P1 (P3, in the strong form): they should be most beneficial precisely where the source-trajectory feedback is unavailable at test time.

\paragraph{Implications for compound AI systems.} This result directly bears on the compound AI systems research community's emphasis on architectural patterns and composition. A Skills layer is one composable component among several (retrieval, tools, verifiers, memory). Our analysis suggests that the marginal value of each component is not independent: a strong tool-grounding layer reduces the marginal value of Skills, and presumably vice versa. Compound-system designers should therefore make these substitution effects explicit in their architectural choices, rather than adding components in additive isolation.

\section{Limitations and Future Work}
\label{sec:limits}

The source study has $N = 15$ challenges, each a multi-phase, long-horizon task averaging 17--20 minutes per run, with three independent trials per condition and a single backbone model. The non-significance result is therefore as much a reflection of statistical power as of effect-size collapse, and we do not claim a causal demonstration of zero effect. What we claim is that \emph{any} effect in this domain is small enough that a 180-run controlled study cannot distinguish it from noise, a striking contrast to the high-power-detected effects in SkillsBench's healthcare and manufacturing domains.

A second alternative explanation we cannot fully rule out is a ceiling effect: the No-Skills baseline already passes 77.8\% of trials, leaving limited headroom for Skills to demonstrate gains. Under this account, Skills might be inert here not because feedback bandwidth substitutes for them but because the tasks are already largely solvable by the model and tool layer alone. Distinguishing the feedback-bandwidth mechanism from a ceiling artifact requires a harder task suite where the No-Skills baseline is substantially lower; the extension we sketch below is designed to address exactly this.

A direct test of H1 would require a SkillsBench-Cyber extension: a larger, verifier-equipped task suite spanning offensive and defensive cybersecurity, evaluated under a $2 \times K$ design that crosses feedback-bandwidth (high MCP-grounding vs.\ raw-shell) with Skills condition. We are working toward such an extension and will release our re-analysis pipeline (scripts, tabulated outcomes, condition mappings) to invite collaboration.

Additionally, the study uses a single backbone model (Claude Sonnet 4.5); extending the ablation to newer frontier and open-source models (e.g., Llama, Mistral, Qwen) will clarify whether the feedback-bandwidth effect is model-family-invariant or depends on instruction-following fidelity.

\section{Conclusion}
\label{sec:conclusion}

We re-analyzed a 180-run controlled study of an MCP ground\-ed autonomous CTF agent as a four-level Skills ablation, and reported a null result at this sample size: $+8.9$~pp marginal benefit, not statistically significant (95\% CI for the difference: $[-6.8, +24.6]$~pp). We proposed the feedback-bandwidth hypothesis as the missing moderator that explains the wide variance in Skills efficacy across domains, and articulated three testable predictions. The result repositions Agent Skills from a universally beneficial procedural-knowledge layer into a compensatory layer whose value is determined by the bandwidth of the environment feedback it sits over.

\bibliographystyle{ACM-Reference-Format}
\bibliography{custom}


\end{document}